\documentclass[5p, twocolumn]{elsarticle}[draft]
\usepackage{amssymb}
\usepackage{amsmath}
\usepackage{diagbox}
\biboptions{numbers,sort&compress}
\usepackage[colorlinks]{hyperref}
\usepackage{caption}
\usepackage[below]{placeins}
\usepackage{color, xcolor}
\usepackage{lineno}
\usepackage{tabularx}
\usepackage{comment}
\usepackage{bbding}
\usepackage{pifont}
\usepackage{multirow}
\let\oldequation\equation
\let\oldendequation\endequation
\renewenvironment{equation}
  {\linenomathNonumbers\oldequation}
  {\oldendequation\endlinenomath}

\begin{document}
\begin{frontmatter}
%\pagewiselinenumbers
%\switchlinenumbers  
\title{Exploring Few-Shot Defect Segmentation in General Industrial Scenarios with Metric Learning and Vision Foundation Models}
\author[Address1]{Tongkun Liu}
\author[Address1,Address2]{Bing Li}
\author[Address1]{Xiao Jin}
\author[Address1]{Yupeng Shi}
\author[Address1]{Qiuying Li}
\author[Address1]{Xiang Wei\corref{cor1}}

\cortext[cor1]{Corresponding author at School of Mechanical Engineering, Xi’an Jiaotong University, Xi'an, Shaanxi, China.}
\address[Address1]{State Key Laboratory for Manufacturing System Engineering, Xi’an Jiaotong University, No.99 Yanxiang Road, Yanta District, 710054, Xi’an, Shaanxi, China}

\address[Address2]{International Joint Research Laboratory for Micro/Nano Manufacturing and Measurement Technologies, Xi’an Jiaotong University, No.99 Yanxiang Road, Yanta District, 710054, Xi’an, Shaanxi, China}

\begin{abstract}
Industrial defect segmentation is critical for manufacturing quality control. Due to the scarcity of training defect samples, few-shot semantic segmentation (FSS) holds significant value in this field. However, existing studies mostly apply FSS to tackle defects on simple textures, without considering more diverse scenarios. This paper aims to address this gap by exploring FSS in broader industrial products with various defect types. To this end, we contribute a new real-world dataset and reorganize some existing datasets to build a more comprehensive few-shot defect segmentation (FDS) benchmark. On this benchmark, we thoroughly investigate metric learning-based FSS methods, including those based on meta-learning and those based on Vision Foundation Models (VFMs). We observe that existing meta-learning-based methods are generally not well-suited for this task, while VFMs hold great potential. We further systematically study the applicability of various VFMs in this task, involving two paradigms: feature matching and the use of Segment Anything (SAM) models. We propose a novel efficient FDS method based on feature matching. Meanwhile, we find that SAM2 is particularly effective for addressing FDS through its video track mode. The contributed dataset and code will be available at: https://github.com/liutongkun/GFDS.   
\end{abstract}

\begin{keyword}
Few-shot Semantic Segmentation, Few-shot Defect Segmentation, Vision Foundation Models
\end{keyword}
\end{frontmatter}

\begin{figure}[t]
    \centering
    \includegraphics[width=\linewidth]{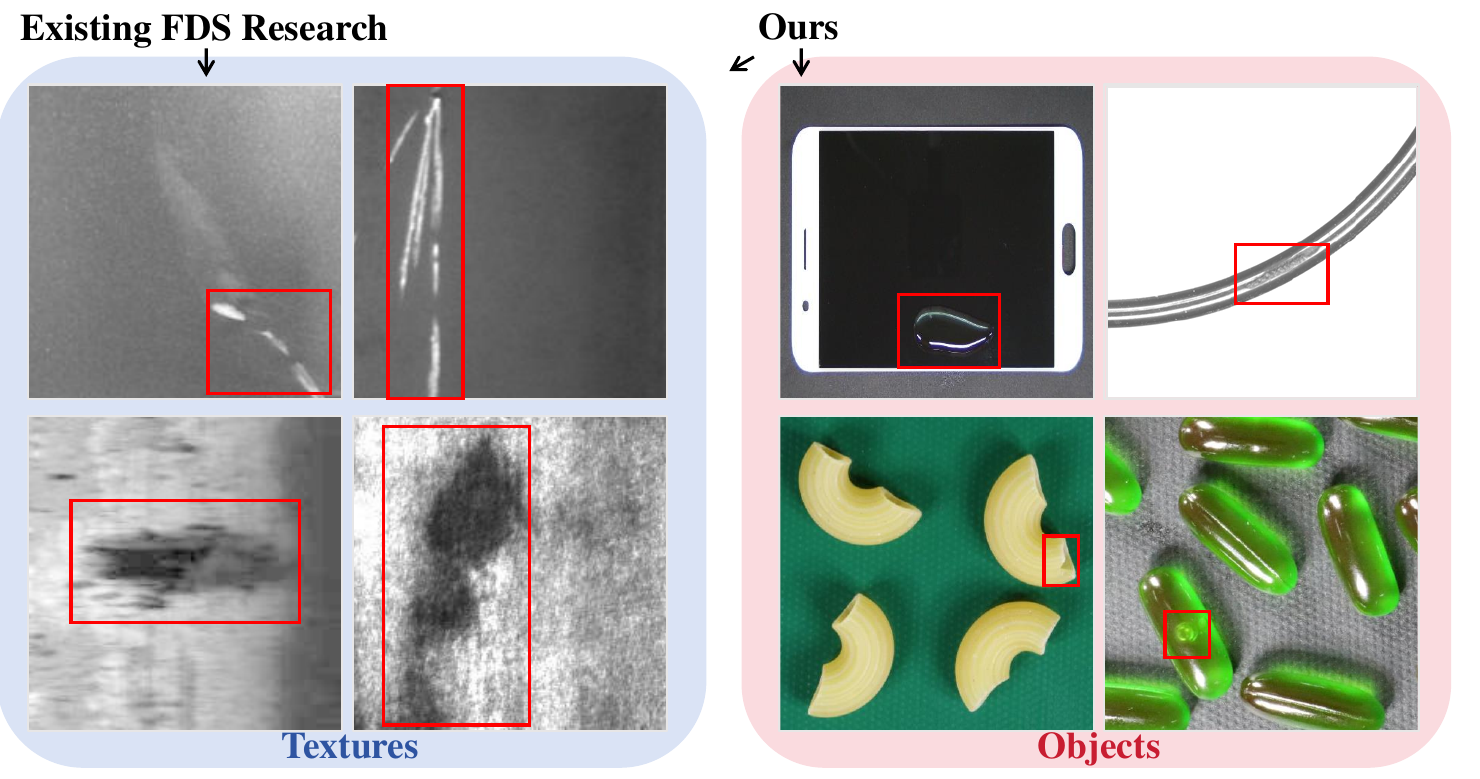}
    \caption{Comparison of existing few shot defect segmentation (FDS) research and ours. The red boxes indicate the segmentation targets. Current FDS research concentrates on textures while ours focuses on more general industrial scenarios. The left part lists four categories of texture defects from the benchmark \cite{bao2021triplet}, which are visibly alike, appearing as white or black spots. This results in a high similarity between the base and test samples in meta-learning.}
 \label{Natural_industrial}
\end{figure}

\section{Introduction}
With the growing emphasis on product quality management in the manufacturing industry, automated industrial visual inspection, which aims to automate the identification of product defects, has become increasingly important. Image semantic segmentation is widely applied in this field since the pixel-level segmentation results can well reflect the location and size of defects, meeting the industrial high precision requirements. However, in industrial applications, traditional supervised segmentation models often face the problem of a shortage of defect samples for training. Compared to common natural images, industrial defective images are typically more difficult to acquire due to commercial privacy and strict production line management processes.

To address the above issue, a possible solution is to introduce few-shot semantic segmentation (FSS) \cite{chang2023few,ren2023visual}. FSS allows for segmenting novel object targets, known as query images, with only a few annotated samples, known as support images. Many FSS methods \cite{tian2020prior, min2021hypercorrelation, lei2022cross, zhang2023personalize, liu2023matcher} adopt metric learning \cite{kaya2019deep}, which achieves segmentation by measuring the similarity between query images and support images in a specific metric space. To find this space, current methods generally follow two paradigms: one involves employing meta-learning \cite{hospedales2021meta}, which trains the model with many additional data; the other utilizes Vision Foundation Models \cite{zhang2023personalize, liu2023matcher} (VFMs), which directly leverages the powerful representations or zero-shot segmentation capabilities of VFMs without fine-tuning.

The above paradigms face several challenges or gaps regarding industrial few-shot defect segmentation (FDS). Specifically, meta-learning-based methods typically require a large number of in-domain training samples. For example, the commonly used FSS benchmarks PASCAL-$5^i$ \cite{xie2021few,shaban2017one} and COCO-$20^i$ \cite{lin2014microsoft} provide 10582 and 82783 images as base samples, respectively. Such quantities are unrealistic for most industrial scenarios. Although some studies have successfully built FDS models \cite{yao2024few,bao2021triplet,yu2022selective} with meta-learning, these methods concentrate on defects in textures like steel, and do not explore more complex industrial contexts, such as objects and multi-component products. Among various industrial products, defects in textures usually have lower inter-class variance, resulting in high similarities between the base samples used in the training phase and the new samples occurring in the testing phase, as shown in the left part of Fig. \ref{Natural_industrial}. Such low inter-class variance may simplify the task and could not demonstrate the effectiveness of meta-learning in more general industrial scenarios. For VFMs-based methods, existing studies mainly focus on natural images \cite{zhang2023personalize, liu2023matcher} or medical images \cite{zhu2024medical, zhao2024retrieval,bai2024fs}, with little exploration of industrial FDS tasks. In particular, industrial production lines often require high efficiency, which is a limitation of many large VFMs. Overall, despite the practical significance of exploring FDS in general scenarios, there is currently a lack of specialized research, either from the meta-learning perspective or the VFM perspective.

\begin{figure}[htbp]
\centering
\includegraphics[width=0.9\linewidth]{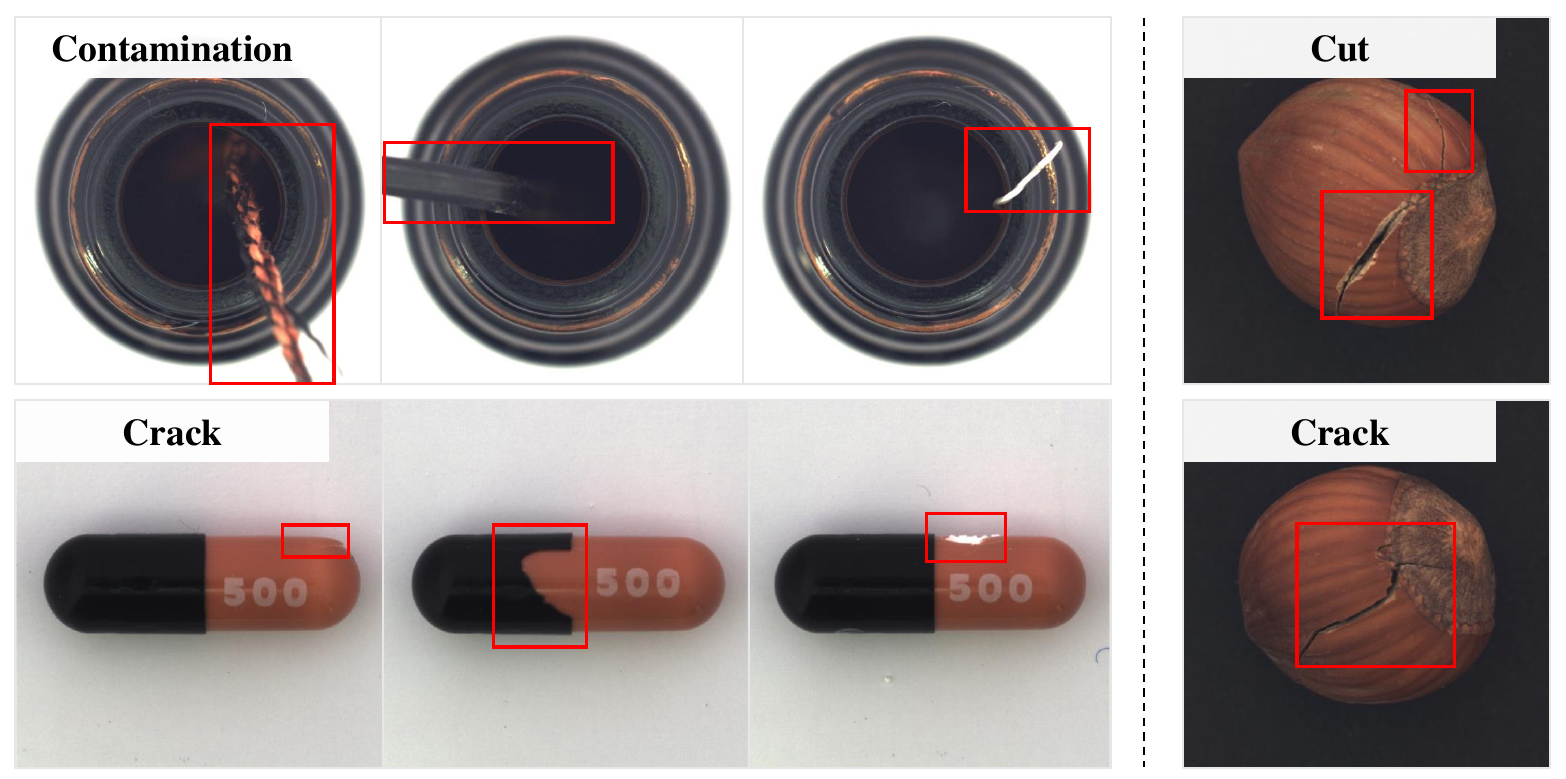}
\caption{Examples of ambiguous defect category definitions from MVTec AD \cite{bergmann2019mvtec}. The defects in the left part exhibit clear pattern differences, yet they are assigned to the same categories. Instead, the defects in the right part appear similar, but they are classified into different categories.}
\label{unclear_definition}
\end{figure}

 In addition to the methodological gaps, we find that publicly available datasets suitable for this task are also imbalanced. Most existing FDS benchmarks \cite{bao2021triplet,yao2024few} include only textures and lack more diverse industrial objects. Although there exist some comprehensive industrial inspection benchmarks \cite{bergmann2019mvtec,zou2022spot}, many of them are specifically designed for anomaly detection, which does not require accurate defect category definitions as those needed for FDS tasks. A typical example is shown in Fig. \ref{unclear_definition}, where the defects in the left part exhibit significant pattern differences but are officially divided into the same defect categories. In contrast, the defects in the right part share highly similar patterns, but are classified into different categories. Such inconsistent standards will pose ambiguity for FDS models. When evaluating the unified performance of FDS on various industrial products, it is essential to ensure that the definition of defect categories maintains a consistent level of semantic abstraction.

Given the aforementioned issues, this paper aims to conduct a comprehensive exploration of metric learning-based FSS methods in general industrial scenarios. Our main work and contributions are summarized as follows:

1. We investigate a valuable yet seldom-explored issue, i.e., FDS in general industrial scenarios. It is worth noting that our `few-shot' refers to only utilizing a small number of defect samples, which is a completely different route from some existing industrial few-shot anomaly detection research \cite{jeong2023winclip,cao2023towards}, where `few-shot' refers to few normal samples. Given the data imbalance issue in existing benchmarks, where texture-based products are more prevalent while object-based products are scarce, we release a new real-world object-based FDS dataset. It contains 9 clearly defined defect categories with pixel-level annotations. Experimentally, we find that the proposed dataset presents more challenges than many existing texture datasets, highlighting the importance of expanding the scope of FDS research to broader industrial scenarios. 
    
2. We provide a more detailed evaluation of classical meta-learning-based FSS paradigms in FDS tasks. Compared to existing research, our novelty lies in: (1) Our evaluation targets are more diverse, beyond just textures. (2) We assess different data utilization strategies and find that they have significant impacts on FDS. Such data-level impacts are often overlooked by existing FDS research, which typically focuses solely on model evaluations.

3. We conduct an in-depth investigation regarding the potential of VFMs in FDS. Specifically, we explore two paradigms: the feature matching-based paradigm, which is generally more efficient, and the SAM \cite{kirillov2023segment} (Segment Anything Model)-based paradigm, which tends to be less efficient but more accurate. For feature matching, our experiments reveal that high-resolution features are crucial for FDS tasks and that appropriate knowledge distillation can effectively enhance performance. Based on these findings, we propose a novel feature-matching method suitable for FDS. For the SAM series models, we fully explore FastSAM \cite{zhao2023fast} and SAM2 \cite{ravi2024sam}. For FastSAM, we design a fusion algorithm to enable its integration with our feature-matching method to enhance performance. For SAM2, we investigate its performance under various prompts and discover that its video track mode is particularly effective for solving FDS.

\section{Related work}
\subsection{Defect segmentation under limited training samples}
The primary challenge of deep learning-based defect segmentation lies in the scarcity of training defect samples. In this context, existing research typically focuses on anomaly detection  \cite{liu2024deep} or FDS \cite{bao2021triplet,yu2022selective,yao2024few}. Anomaly detection is well-suited for scenarios with abundant normal samples. It has received significant attention in recent years, including the development of comprehensive benchmarks \cite{bergmann2019mvtec, bergmann2021mvtec, zou2022spot}, research on unified models \cite{10574313,you2022unified}, and the ongoing exploration of new vision foundation models \cite{roth2022towards,jeong2023winclip,cao2023towards,li2024clipsam}. FDS differs from anomaly detection in that it aims to utilize a small number of defect samples, rather than normal samples, to perform defect segmentation, thus avoiding the need to collect numerous normal samples. It also holds significant practical value. However, current research on FDS is generally less comprehensive, with most benchmarks \cite{bao2021triplet,yao2024few} and methods concentrating on texture-based scenarios and limited exploration of VFMs. Our research aims to investigate FDS in more general industrial scenarios, contributing both to datasets and the exploration of metric learning-based methods.

\subsection{Few-shot semantic segmentation}
FSS can be generally achieved through fine-tuning \cite{nakamura2019revisiting} or metric learning. Fine-tuning-based methods leverage the few available labeled samples to optimize the pre-trained model weights, whereas for metric learning-based methods, these few samples usually only appear during the testing phase for evaluating the similarities. Compared to fine-tuning-based methods, metric learning-based methods often offer greater value in industrial scenarios because they can continuously utilize new defect samples generated on the production line without the need to retrain the model. Therefore, our study focuses on metric learning-based methods, and we broadly classify them into two categories: meta-learning-based methods and VFMs-based methods.

\noindent \textbf{Meta-learning-based methods.} Meta-learning typically refers to leveraging auxiliary data and tasks to improve the model's learning capability, also known as `learning to learn'. In metric learning, the goal of meta-learning is to find a metric space where samples from the same class are close to each other, while samples from different classes are distant. Common methods include using single \cite{dong2018few,fan2022self,tian2020prior}, multiple prototypes \cite{li2021adaptive}, and employing pixel-to-pixel dense correspondences \cite{min2021hypercorrelation}. These methods are primarily evaluated on natural image datasets such as PASCAL-$5^i$ and COCO-$20^i$, which can provide numerous base samples from the same domain for training. However, in many applications, collecting sufficient base samples is impractical. In this context, some research introduces cross-domain few-shot semantic segmentation (CD-FSS) \cite{guo2020broader, lei2022cross}, which aims to generalize the knowledge learned from domains with sufficient base samples to new low-resource domains. Existing CD-FSS research primarily focuses on the cross-domain issues between different natural image datasets \cite{li2020fss}, medical image datasets \cite{candemir2013lung,jaeger2013automatic}, and satellite image datasets \cite{demir2018deepglobe}, with limited attention to the industrial domain. Some studies \cite{bao2021triplet,yu2022selective,yao2024few} introduce meta-learning in industrial texture scenarios, such as steel and metal surfaces. However, the defects in these scenarios often exhibit high similarity, which can be less challenging compared to general industrial scenarios. To fill this gap, our research comprehensively evaluates several classic meta-learning methods for general FDS tasks. In particular, our evaluations additionally consider the impact of different training data strategies. We observe that the impact is significant but is often overlooked in existing FDS research.

\begin{figure*}[tbp]
    \centering
    \includegraphics[width=\linewidth]{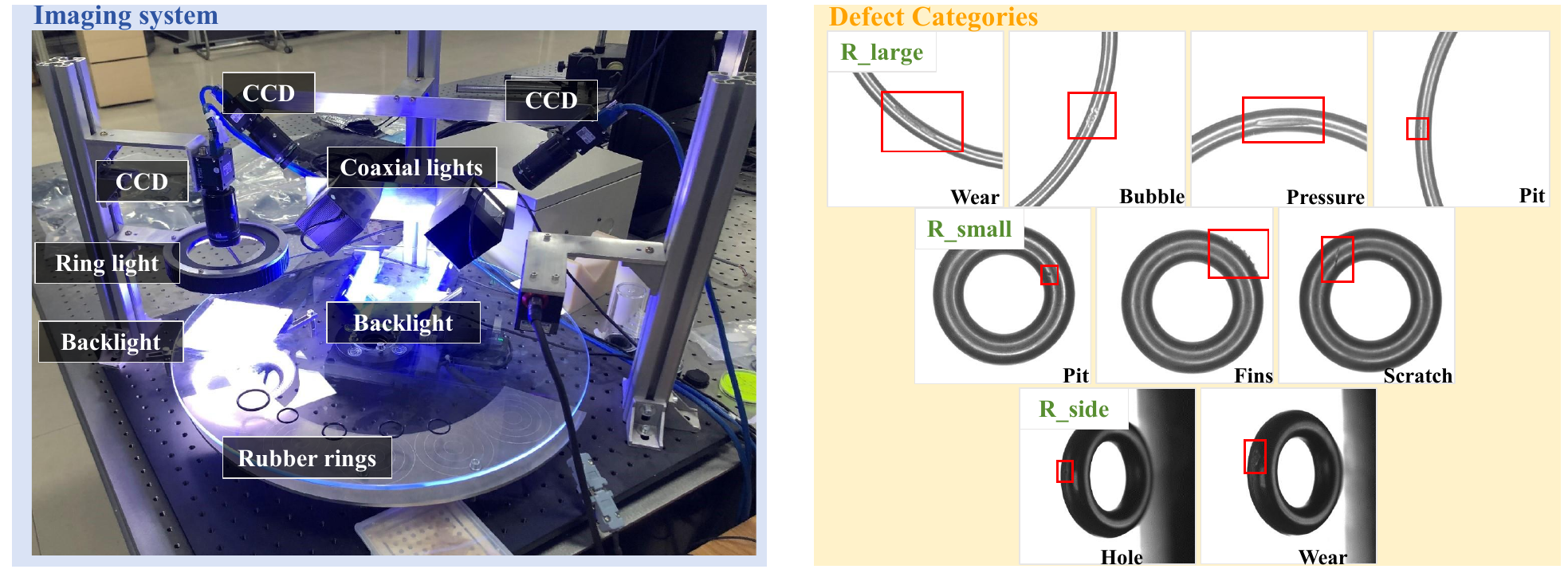}
	\caption{Our contributed dataset. The proposed dataset contains three types of rubber ring images: large rubber rings, small rubber rings, and side views of rubber rings, abbreviated as `R\_large', `R\_small', and `R\_side', respectively. They contain a total of nine types of defects.}
	\label{rubberring_dataset}
\end{figure*}

\noindent \textbf{VFMs-based methods.}
Vision foundation models refer to large-scale pre-trained models that can serve as a base for various vision tasks. These models can extract generalized features, exhibit strong transfer capabilities, or even directly perform zero-shot inference. Early VFMs are primarily based on convolutional neural network (CNN) architectures, such as the ImageNet\cite{deng2009imagenet}-supervised pre-trained ResNet \cite{he2016deep} models. These models often serve as backbones for many FSS methods. Subsequently, vision transformer (ViT) \cite{alexey2020image} models trained with self-supervised learning, such as DINO \cite{caron2021emerging} and DINOv2 \cite{oquab2023dinov2}, demonstrate powerful representation learning capabilities, allowing image semantic segmentation through simple KNN methods. Recently, Segment Anything Models (SAM) \cite{kirillov2023segment,ravi2024sam,zhao2023fast,zhang2023faster,xiong2024efficientsam,songa2024sam} have emerged, capable of achieving convincing zero-shot image segmentation on various datasets with prompt engineering. Some studies apply SAM series models to FSS tasks in natural images \cite{zhang2023personalize,liu2023matcher} and medical images \cite{zhu2024medical, zhao2024retrieval,bai2024fs}, investigating domain-specific prompts or adaptation strategies. Yet, there is currently a lack of exploration regarding industrial fields. Our research fills this gap by exploring VFMs in FDS from two aspects, including feature matching and the use of SAM models.

\begin{table*}[htbp]
\centering
\caption{Details of the benchmark. Each sub-dataset in the table represents a single type of industrial product, except for DAGM. DAGM contains ten different products, each with only one type of defect.}
\scalebox{0.52}{
\begin{tabular}{c|ccccccccccccccc}
\hline
Dataset                                                                       & Tile      & Grid       & Steel  & Kol       & \multicolumn{2}{c}{PSPs} & \multicolumn{2}{c}{DAGM} & R\_small    & R\_large     & R\_side  & MSD          & Capsules     & \multicolumn{2}{c}{Macaroni2} \\ \hline
\multirow{5}{*}{\begin{tabular}[c]{@{}c@{}}Category\\ (quantity)\end{tabular}} & Crack(11) & Bent(12)   & Am(50) & Crack(52) & Liquid(19)   & Poke(24)  & C1(71)     & C2(84)      & Scratch(30) & Wear(11)     & Hole(16) & Oil(400)     & Bubble(30)   & Break(10)      & Color1(20)   \\
                                                                               & Gray(16)  & Glue(11)   & Sc(50) & -         & Mark(14)     & Water(25) & C3(84)     & C4(68)      & Fins(15)    & Pressure(19) & Wear(11) & Scratch(400) & Discolor(15) & Color2(12)     & Scratch(15)    \\
                                                                               & Rough(15) & Thread(11) & Ld(48) & -         & Oil(16)      & Spot(35)  & C5(80)     & C6(67)      & Pit(36)     & Bubble(26)   & -        & Stain(400)   & Leak(20)     & Edge(25)       & Crack(15)    \\
                                                                               & Glue(18)  & Broken(12) & -      & -         & Scratch(16)  & -         & C7(150)    & C8(150)     & -           & Pit(33)      & -        & -            & Misshape(20) & -    & -            \\
                                                                               & Oil(18)   & Metal(11)  & -      & -         & -            & -         & C9(150)    & C10(150)    & -           & -            & -        & -            & Scratch(15)  & -              & -            \\ \hline
\begin{tabular}[c]{@{}c@{}}Total quantity\end{tabular}                      & 78        & 57         & 148    & 52        & \multicolumn{2}{c}{149}  & \multicolumn{2}{c}{1054} & 81          & 89           & 27       & 1200         & 100          & \multicolumn{2}{c}{97}       \\ \hline
\end{tabular}}
\label{Table_benchmark}
\end{table*}

\section{The comprehensive FDS benchmark}
\label{Sec_proposed_benchmark}
\subsection{The contributed dataset}
Our contributed dataset is derived from real-world rubber ring defect detection cases. Rubber rings primarily function to enhance sealing, preventing the leakage of gases and liquids, or to provide vibration damping. They are widely used in various industries, such as automotive, aerospace, medical devices, and consumer electronics. These applications require the rubber ring to be of high quality, as the defects can compromise their air-tightness or strengths, leading to product failure or reduced performance. The proposed dataset contains nine types of rubber ring defects in total. Fig. \ref{rubberring_dataset} shows our data collection platform and the defect examples. Table. \ref{Table_benchmark} provides detailed information on their quantities.

\subsection{The reorganization of existing datasets}
\begin{figure*}[thp]
    \centering
    \includegraphics[width=\linewidth]{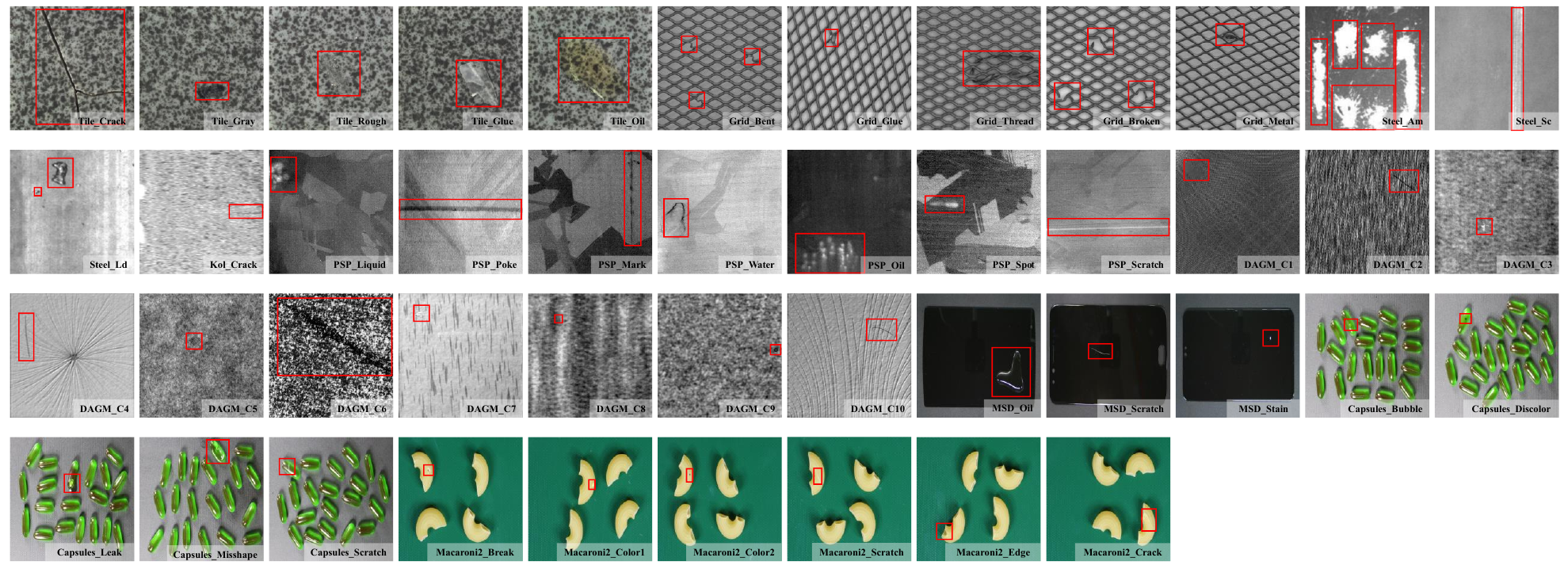}
	\caption{Examples of different product defects selected from existing publicly available datasets.}
	\label{benchmark_figure}
\end{figure*}

We select several defect segmentation datasets from existing publicly available benchmarks. Our selection criteria are that the datasets must have clearly defined defect categories suitable for FDS tasks and should cover a wide range of industrial scenarios. The selected datasets involve textures, single-component objects, and multi-component objects. Specifically, textures include `tile' and `grid' from \cite{bergmann2019mvtec}, `steel' from \cite{bao2021triplet}, defected electrical commutator (referred to as KolektorSDD) from \cite{tabernik2020segmentation}, polycrystalline silicon panels (PSPs) from \cite{yao2024few} and synthetic texture defects (referred to as DAGM) from \cite{wieler2007weakly}. In particular, we re-annotate the coarse labels originally provided in DAGM to make them suitable for evaluating defect segmentation performance. Single-component objects include mobile phone screen defects (MSD) from \cite{zhang2022fdsnet} and our contributed rubber ring dataset. Multi-component objects include `capsules' and `macaroni2' from \cite{zou2022spot}. Fig. \ref{benchmark_figure} exhibits some examples from the selected datasets. The quantitative details are shown in Table. \ref{Table_benchmark}.

\section{Problem Setting}
The testing set $\mathcal{D}_{\rm test}$ in FSS is composed of the support set $\mathcal{S} = (I_{s}, M_{s})$ and the query set $\mathcal{Q} = (I_{q}, M_{q})$, where $I_{*}$ represents the image and $M_{*}$ represents the corresponding annotation mask. In the testing phase, $M_{q}$ is unknown and the goal of the task is to utilize $I_{s}$, $M_{s}$, and $I_{q}$ to obtain $M_{q}$. If the support set $\mathcal{S}$ contains $K$ categories, with $N$ samples per category, this setting is referred to as `$K$-way $N$-shot'. FSS typically allows the inclusion of an additional training set $\mathcal{D}_{\rm train}$. The annotated object categories in $\mathcal{D}_{\rm train}$ must have no overlap with the target categories in $\mathcal{D}_{\rm test}$. Therefore, $\mathcal{D}_{\rm test}$ is also referred to as the novel set, indicating that its object categories have not been seen during the training phase, while $\mathcal{D}_{\rm train}$ is also referred to as the base set. On the other hand, although $\mathcal{D}_{\rm train}$ and $\mathcal{D}_{\rm test}$ consists of different object categories, conventional meta-learning-based FSS task requires that $\mathcal{D}_{\rm train}$ and $\mathcal{D}_{\rm test}$ are from the same domain. When $\mathcal{D}_{\rm train}$ and $\mathcal{D}_{\rm test}$ come from different domains, it is termed cross-domain few-shot semantic segmentation.

\section{Method}
\label{Sec_method}
\begin{figure*}[htbp]
    \centering
    \includegraphics[width=\linewidth]{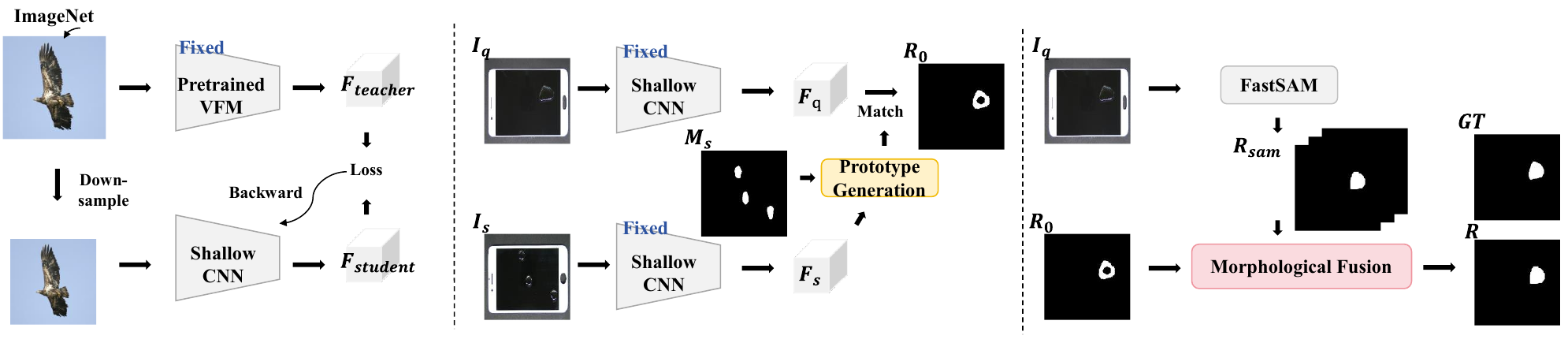}
	\caption{The overview of the proposed feature matching-based FDS method. It primarily consists of three parts: 1. Feature distillation, 2. Feature matching, and 3. Refining the results with FastSAM.}
	\label{methodoverview}
\end{figure*}

In this section, we will illustrate how to construct FDS methods using VFMs. We propose a novel and efficient FDS method based on feature matching (Fig. \ref{methodoverview}), described in Sec. \ref{proposed_method}. Besides, we find that the video track mode of SAM2 can be directly used to achieve accurate FDS, described in Sec. \ref{Exploring SAM2}.

\subsection{The proposed method}
\label{proposed_method}
\subsubsection{Feature distillation}
Some powerful pre-trained model\cite{caron2021emerging,oquab2023dinov2} features have been proven to contain explicit information for image segmentation. Therefore, an intuitive approach for FDS is to directly leverage these powerful pre-trained features for feature matching. Following this paradigm, the first issue lies in selecting the appropriate pre-trained features. Experimentally, we find that representing industrial defects typically requires features with high resolution, probably because industrial defects are usually small in size. This can pose a challenge for existing pre-trained models. Specifically, for ViT models, high-resolution features generally require smaller patch sizes, which significantly increase computational overhead. For CNN models, high-resolution features correspond to features from shallow layers, which often lack sufficient representation abilities. To address this issue, we here introduce knowledge distillation. We adopt a pre-trained ViT model with a small patch size as the teacher model and a shallow CNN network as the student model. The student model is required to mimic the outputs of the teacher model. In this context, the features output by the shallow CNN model are high-resolution, and through knowledge distillation, it holds the potential to acquire the high-semantic features from the ViT model while maintaining a rapid inference speed. In particular, the input image resolution of the teacher model is set to be higher to ensure that its features align spatially with the outputs of the student model. To implement this distillation process, we introduce the ImageNet dataset as the training set. The distilled student model will then serve as our feature extractor, denoted as $E(\cdot)$ in the subsequent section.

\subsubsection{Feature matching}
Given the feature extractor $E(\cdot)$, we use it to extract the features of the support and query images as
\begin{equation}
F_s = E(I_s), \text{ } F_q = E(I_q),
\end{equation}
where $I_s, I_q \in \mathbb{R}^{H \times W \times C}$. $H, W$ are the image's spatial dimensions and $C$ is the channel number. $F_s, F_q  \in \mathbb{R}^{H^{'} \times W^{'} \times C^{'}}$ are the corresponding feature maps with the spatial dimensions and channel number of $H^{'}, W^{'}, C^{'}$. Then, we downsample the binary support mask $M_s \in \mathbb{R}^{H \times W \times 1}$ to $M_{s}^{'} \in \mathbb{R}^{H^{'} \times W^{'} \times 1}$. We utilize $M_{s}^{'}$ to partition $F_s$ into foreground  vectors $F_s^f \in \mathbb{R}^{C^{'}\times n_f}$ and background vectors $F_s^b \in \mathbb{R}^{C^{'}\times n_b}$, where $n_f$ and $n_b$ are the number of feature vectors. The process is formulated as
\begin{equation}
    F_s^f = \{f_{ij} \mid f_{ij} \in F_s \text{ and } M_{s\_ij}^{'}>0\}
\end{equation}

\begin{equation}
    F_s^b = \{f_{ij} \mid f_{ij} \in F_s \text{ and } M_{s\_ij}^{'}=0\},  
\end{equation}
where $i,j$ represent the spatial coordinates. The crux of feature matching lies in the construction of appropriate prototypes utilizing $F_s^f$ and $F_s^b$. Many learnable FSS methods \cite{fan2022self, tian2020prior} construct prototypes using global pooling. In our task, this approach is suboptimal for two principal reasons: first, we do not have an additional learning process to adjust the prototypes, which results in significant information loss when global pooling is applied directly; second, many industrial components are fine-grained, making the global pooling operation inherently unsuitable. Empirically, we directly retain $F_s^b$ without any adjustment as the background prototype. For $F_s^f$, we aggregate it into patches using patch averaging to reduce the impact of noise. Our final foreground prototype is written as
\begin{equation}
    F_s^{f_{patch}} = PatchAvg(F_s^f), \text{ } F_s^{f_{patch}} \in \mathbb{R}^{C^{'}\times n_f}
\end{equation}
where $PatchAvg(\cdot)$ denotes the patch averaging, and the patch size we use is 3 with a stride of 1. Given the query feature $F_q$, we compute the cosine similarities of $F_q$ with respect to $F_s^{f_{patch}}$ and $F_s^b$ as

\begin{equation}
    F_q^f = \frac{F_q \cdot F_s^{f_{patch}}}{\|F_q\|_2 \|F_s^{f_{patch}}\|_2} 
\end{equation}

\begin{equation}
    F_q^b = \frac{F_q \cdot F_s^b}{\|F_q\|_2 \|F_s^b\|_2}
\end{equation}
where $F_q^f \in \mathbb{R}^{H^{'} \times W^{'} \times n_f}$ and $F_q^b \in \mathbb{R}^{H^{'} \times W^{'} \times n_b}$. We take the maximum value of $F_q^f$ and $F_q^b$ along the last channel dimension to obtain the foreground and background similarity maps $F_q^{f*} \in \mathbb{R}^{H^{'} \times W^{'}}$ and $F_q^{b*} \in \mathbb{R}^{H^{'} \times W^{'}}$ as
\begin{equation}
    F_q^{f*}(i,j) = \max_{k \in \{1, 2, \dots, n_f\}} F_q^{f}(i,j,k)
\end{equation}
and 
\begin{equation}
F_q^{b*}(i,j) = \max_{k \in \{1, 2, \dots, n_b\}} F_q^{b}(i,j,k).
\end{equation}
We concatenate $F_q^{b*}$ and $F_q^{b*}$, apply the `argmax' operation along the last channel dimension, and then upsample the result to the original image size to obtain the segmentation result $R_0 \in \mathbb{R}^{H \times W}$.

\begin{figure}[htbp]
    \centering	
    \includegraphics[width=\linewidth]{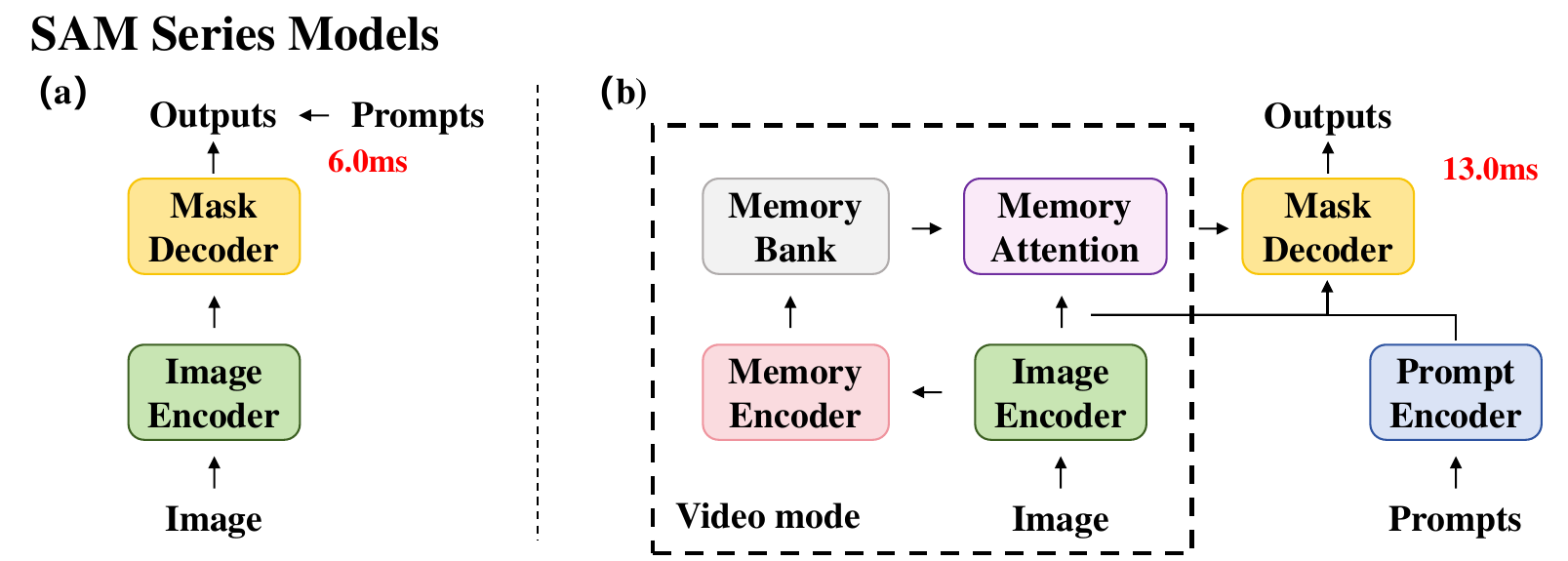}
	\caption{The architectures of the SAM models. (a). FastSAM; (b). SAM2. We mark their inference times in our environment (described in Sec. \ref{evaluateenv}) in red. For SAM2, we test the inference time of its video track mode with the official small model.}
	\label{SAM_ilu}
\end{figure}

\subsubsection{Refining Feature Matching with FastSAM}
$R_0$ can be further refined using the SAM series models. Considering the inference efficiency, we here choose FastSAM \cite{zhao2023fast}. FastSAM is a CNN-based YOLO \cite{jocher2023yolo} model that can be used for zero-shot image segmentation. It differs from other ViT-based SAM models \cite{kirillov2023segment, ravi2024sam, zhang2023faster, xiong2024efficientsam,songa2024sam} in that it directly generates all the potential segmentation masks in a single forward pass, without requiring the decoder to process various pre-encoded prompts, as shown in Fig. \ref{SAM_ilu}, (a). We first input $F_q$ into FastSAM to perform zero-shot segmentation. We eliminate the overlapping regions among the segmented masks and denote the resulting set as $M_{\rm sam}$, which can be formulated as

\begin{equation}
    M_{\rm sam} = \{ m_1, m_2, \dots m_n \},
\end{equation}
and
\begin{equation}
    \left( \bigcup_{i=1}^{n} m_i = I \right) \quad \text{and} \quad (m_i \cap m_j = \emptyset \quad \text{for} \quad i \neq j),
\end{equation}
where $m_i \in \mathbb{R}^{H \times W}$. $I \in \mathbb{R}^{H \times W}$ refers to the identity matrix. $n$ is the number of the masks. We then use $R_0$ to filter the masks in $M_{\rm sam}$, selecting those masks that are largely covered by $R_0$. The process can be formulated as   
\begin{equation}
\label{eq_tau1}
M_{\rm select} = \{m_{i} \mid m_i \in M_{\rm sam} \text{ and } \frac{m_{i} \cdot R_0}{m_{i}}>\tau_1 \},
\end{equation}
where $\tau_1$ is a hyperparameter for the overlap threshold and we set it as 0.2. The segmentation result obtained from FastSAM can be expressed as 
\begin{equation}
    R_{\rm sam}=\bigcup_{m_i \in M_{\rm select}} m_i.
\end{equation}

Existing methods \cite{liu2023matcher,zhang2023personalize} tend to only use $R_{\rm sam}$ as the final output, but in our task, we find it usually suboptimal since FastSAM often overlooks small or less noticeable defects. Another intuitive approach is to directly perform a union operation on $R_0$ and $R_{\rm sam}$. This allows $R_0$ and $R_{\rm sam}$ to complement each other's incomplete segmentation, but it is ineffective in eliminating false positives in $R_0$. To address these issues, we design a more specialized fusion approach based on morphological operations. We perform morphological connected component analysis on $R_0$, dividing it into multiple disconnected regions $\Re_0 =\{R_0^1, R_0^2, \cdots R_0^{n_r}\}$. We also apply the morphological dilation operation $Dil(\cdot)$ to each mask in $M_{\rm select}$ and denote the dilated mask as $Dil(m_{i})$. The dilation operation here is used to mitigate the impact of misalignment, and we set the dilation kernel size to 21. For any $R_0^l \in \Re_0$, we observe that if there exists a $m_i \in M_{\rm select}$ such that $R_0^l$ is sufficiently covered by $Dil(m_{i})$, it typically indicates that $R_0^l$ and $m_i$ correspond to the same defect region. At this time, since $m_{i}$ generally provides more refined boundaries compared to feature matching, $R_0^l$ should be replaced with $m_i$.  Conversely, if there is no sufficient coverage, it indicates that $R_0^l$ and $m_i$ represent different defect regions, and both should be retained. We denote the retained $\Re_0$ as $\Re_1$, which can be written as
\begin{align}
\Re_1 &= \{ R_0^l \mid R_0^l \in \Re_0 \text{ and } \nonumber \\
&\quad \forall m_{i} \in M_{\rm select}, \frac{Dil(m_{i}) \cdot R_0^l}{R_0^l} < \tau_2 \}
\end{align} where $\tau_2$ is the hyperparameter to control the overlap threshold and we set it as 0.9. The final segmentation result $R$ is formulated as
\begin{equation}
R = R_{\rm sam} \cup \left( \bigcup_{R_0^l \in \Re_1} R_0^l \right).
\end{equation}

\subsection{Exploring SAM2 in FDS}
\label{Exploring SAM2}
Our proposed method addresses FDS with feature matching, which generally offers more simplicity and efficiency. Regarding other VFMs, we also find that SAM2 \cite{ravi2024sam} can be directly applied in FDS by utilizing its video track mode. The basic architecture of SAM2 is shown in Fig. \ref{SAM_ilu}, (b). SAM2 includes three structures as the original SAM \cite{kirillov2023segment}: the image encoder, prompt encoder, and mask decoder. With these components, SAM2 can perform prompt-based image segmentation. Specifically, the image encoder encodes the input image, while the prompt encoder encodes the input prompts, such as points, boxes, or masks. Both encoded embeddings are then fed into the mask decoder to generate the final output. Besides, SAM2 introduces an additional memory mechanism, which is originally designed for video track tasks. Features from the previous frame are encoded by the memory encoder and stored in the memory bank. These features are then used to compute memory attention with the current frame's features, which are subsequently passed into the mask decoder to track the target in the current frame.

For FDS, SAM2 can also be used to refine the results obtained from feature matching. However, through experimentation, we find it suboptimal. A more direct and effective approach is to treat FDS as a video track task, where the support set is considered as the previous frame and the query set as the current frame. SAM2’s video track mode can then be used to directly perform this task. To implement this process, SAM2's image encoder $E_{img}(\cdot)$ is used to extract features from $I_q$ and $I_s$, denoted as $f_q$ and $f_s$ respectively. Then, the memory encoder $E_{mem}(\cdot)$ further encodes $f_s$ and $M_s$, resulting in the memory feature $f_{mem}$, which can be written as:
\begin{equation}
f_{mem} = E_{mem}(f_s, M_s).
\end{equation}
$f_{mem}$ is used to compute memory attention with $f_q$, and the attention result is fed into the mask decoder to obtain the final segmentation result $R_{sam2}$. The process can be written as:
\begin{equation}
R_{sam2} = D_{mask}(Atten(f_q, f_{mem})),
\end{equation}
where $Atten(\cdot)$ represents the memory attention and $D_{mask}$ denotes the mask decoder.

\section{Experiment}
\label{Sec_Experiment}
\subsection{The selected methods}
In this section, we evaluate meta-learning-based methods and VFMs-based methods on the comprehensive FDS benchmark proposed in Sec. \ref{Sec_proposed_benchmark}. We resize the images in the benchmark uniformly to $256 \times 256$. Basically, we implement the selected methods with their official codes but slightly modified the downsampling operation of the support mask. Specifically, we observe that the commonly used bilinear interpolation for downsampling leads to small defects being overlooked in our benchmark. Therefore, when these codes involve this operation, we add an additional condition to switch the bilinear interpolation to nearest neighbor downsampling if the defect size is too small. The remaining details are presented as follows.

\subsubsection{Meta-learning-based methods.}
The selected methods include single-domain approaches PFENet \cite{tian2020prior}, SSP \cite{fan2022self}, and HSNet \cite{min2021hypercorrelation} and the cross-domain approach PATNet \cite{lei2022cross}. We select ImageNet-pretrained ResNet-50 as the backbone for these methods. During the training phase, we utilize random flipping and 90-degree rotations as data augmentations. To explore the impact of training data on the FDS task, we employ two data usage strategies: (1). \textbf{Cross-domain setting}: Each industrial product can be viewed as a distinct domain. When there is no overlap between the products contained in the training set and the testing set, it is considered a cross-domain setting. Here, we divide the entire benchmark into two folds for cross-validation, as shown in Table. \ref{Cross_Fold_Table}. (2). \textbf{In-domain setting}: This refers to the training set and testing set containing the same industrial products. Under this setting, we select one defect class from each product as the testing set, and the remaining defect classes from the same product are used as the training set. In particular, for Kol and DAGM, since each of their products only contains one type of defect, we do not perform in-domain validation on them. Empirically, we train PFENet for 400 epochs, SSP for 200 epochs, and both HSNet and PATNet for 800 epochs in both the two settings, using the default optimization strategies of their official codes.

\begin{table}[htbp]
\centering
\caption{Fold division for cross-domain setting evaluation. The `Quantity' column consists of defect category quantities and image quantities, represented as `categories/images'}
\scalebox{0.7}{
\begin{tabular}{c|c|c}
\hline
  & Category                                          & Quantity \\ \hline
Fold0 & DAGM, R\_large, R\_side, R\_small, Capsules, Tile & 29/1429  \\
Fold1 & Phone, Grid, Steel, Kol, PSP, Macaroni2           & 25/1703  \\ \hline
\end{tabular}}
\label{Cross_Fold_Table}
\end{table}

\subsubsection{VFMs-based methods.}
We evaluate PerSAM \cite{zhang2023personalize}, Matcher \cite{liu2023matcher}, the video track mode of SAM2 \cite{ravi2024sam}, and our proposed feature matching method.

\noindent \textbf{PerSAM and Matcher.} Both these methods adopt SAM \cite{kirillov2023segment} with ViT-H by default. For the feature extractors, PerSAM adopts the SAM's image encoder and Matcher introduces DINOv2 \cite{oquab2023dinov2} with ViT-L/14. For Matcher, we upsample the image to $266 \times 266$ during the feature extraction phase to ensure it is divisible by the corresponding patch size of its feature extractor. Meanwhile, the SAM model used in PerSAM and Matcher requires the image to be interpolated to $1024 \times 1024$ to ensure the alignment of positional encoding. 

\noindent \textbf{The video track mode of SAM2.} We evaluate the different model weights provided by SAM2.1, including its large model, base model, small model, and tiny model, denoted as SAM2-l, SAM2-b, SAM2-s, and SAM2-t respectively. SAM2 is officially trained on images with a resolution of $1024 \times 1024$, but the official video mode allows scaling of positional encodings to adapt to different image resolutions. Therefore, we also conduct multiple experiments by using the original images and upsampling them to resolutions of $512 \times 512$ and $1024 \times 1024$.

\noindent \textbf{Feature matching.} In the feature distillation process, we use DINO \cite{caron2021emerging} with ViT-S/8 \cite{alexey2020image} as the teacher model and select the patch features from the last transformer block as the output. For the student model, we use the shallow CNN network proposed in \cite{Batzner_2024_WACV}, which is composed of some simple convolutional and pooling layers, as shown in Table. \ref{studentmodel}. We try two student model widths, i.e., $c=256$ and $c=128$, denoted as FM-l and FM-s respectively. We train these models on the ImageNet dataset for 160000 iterations with a batch size of 12. The input image for the teacher model is resized to $512 \times 512$ and for the student models, it is resized to $256 \times 256$. We use the $l_2$ loss and the Adam optimizer and set the learning rate as 1e-4 and the weight decay as 1e-5. Regarding FastSAM, we evaluate both the default model and its lightweight version, FastSAM-s. For these FastSAM models, we use the default hyperparameters except for adjusting `iou' and `confidence' to 0.5 and 0.1 respectively.      

\begin{table}[htbp]
\centering
\caption{The architecture of the student network. `in' and `out' refer to the input and output channel numbers respectively. `c' refers to the base channel number and we set it as 256 for FM-l and 128 for FM-s. `Conv2D' and `AvgPool2D' refer to the 2D convolutional layer and the average pooling layer respectively.}
\scalebox{0.7}{
\begin{tabular}{c|ccccc}
\hline
            & in & out & kernel size & stride & padding \\ \hline
Conv2D+ReLu & 3          & c           & 4           & 1      & 3       \\
AvgPool2D   & c          & c           & 2           & 2      & 1       \\
Conv2D+ReLu & c          & 2c          & 4           & 1      & 3       \\
AvgPool2D   & 2c         & 2c          & 2           & 2      & 1       \\
Conv2D+ReLu & 2c         & 2c          & 3           & 1      & 1       \\
Conv2D      & 2c         & 384         & 4           & 1      & 0       \\ \hline
\end{tabular}}
\label{studentmodel}
\end{table}

\subsection{Evaluation metrics.}
We select two commonly used metrics in FSS including mean intersection over union (mIoU) and foreground-background IoU (FB-IoU). Higher values of these metrics indicate better model performance. Their formulas are:
\begin{equation}
    \rm mIoU = \frac{1}{C} \sum_{c=1}^{C} \rm IoU_{c}
\end{equation}

\begin{equation}
    \rm FB \text{-} IoU = \frac{1}{2} (\rm IoU_{F}+IoU_{B})
\end{equation}
where $C$ is the total number of target classes in the test set, excluding the background class. $\rm IoU_{c}$ represents the intersection over union of class $c$. FB-IoU treats all target classes as a unified foreground class and considers only the foreground and background IoU, denoted as $\rm IoU_{F}$ and $\rm IoU_{B}$, respectively.

\subsection{Implementation Environment.}
\label{evaluateenv}
The experiments are conducted in Ubuntu 20.04 using Pytorch 2.5.1 with an NVIDIA GeForce GTX 3090Ti and I7-12700. For the inference speed test, we set the batch size to 1, use the mixed precision mode of Pytorch as in the official SAM2 code, and exclude the processing time of the support image. 

\begin{table*}[htbp]
\centering
\caption{mIoU and FB-IoU results (written as `mIoU, FB-IoU') on the proposed benchmark under `1-shot 1-way' setting. Methods marked with `*' indicate adopting the `in-domain' training paradigm. `FPS' refers to `frames per second'. Kol and DAGM are only evaluated on the cross-domain setting since each of their products only contains one type of defect. In the `Average' column, mIoU represents the average value across 54 defect categories, while FB-IoU represents the average value across 12 product types. For SAM2, we specify the image interpolation resolution in parentheses.}
\scalebox{0.55}{
\begin{tabular}{c|cccccccccccc|c|c}
\hline
Method        & Tile       & Grid       & Steel      & Kol        & PSPs       & DAGM       & R\_small   & R\_large   & R\_side    & MSD        & Capsules   & Macaroni2  & Average    & FPS   \\ \hline
PFENet        & 0.0, 45.0  & 0.0, 49.5  & 17.3, 51.0 & 10.6, 55.0 & 2.0, 50.0  & 0.0, 49.6  & 0.0, 49.9  & 0.0, 49.6  & 0.0, 49.9  & 12.3, 59.4 & 0.0, 49.7  & 0.0, 45.0  & 2.1, 50.3  & 154.0 \\
PFENet*       & 56.5, 76.2 & 28.0, 63.5 & 43.6, 68.2 & \textcolor{gray}{10.6, 55.0} & 23.2, 60.9 & \textcolor{gray}{0.0, 49.6}  & 0.0, 49.9  & 28.3, 64.0 & 0.0, 49.9  & 7.4, 53.2  & \textbf{24.0, 62.1} & 0.0, 50.0  & 18.6, 58.6 & 154.0 \\
SSP           & 54.1, 74.4 & 29.4, 63.2 & 50.4, 71.0 & 2.2, 43.7  & 16.7, 55.8 & 21.2, 59.5 & 0.6, 40.6  & 6.2, 47.3  & 1.2, 40.3  & 22.7, 60.7 & 2.4, 44.3  & 1.0, 47.6  & 18.8, 54.0 & 302.4 \\
SSP*          & 65.3, 80.5 & 38.0, 68.4 & 61.2, 77.6 & \textcolor{gray}{2.2, 43.7}  & 34.5, 66.3 & \textcolor{gray}{21.2, 59.5} & 10.2, 54.7 & 33.4, 65.9 & 34.1, 65.7 & 26.8, 62.8 & 23.1, \textbf{62.1} & 9.4, 53.8  & 30.4 63.4  & 302.4 \\
HSNet         & 7.1, 47.4  & 2.6, 50.4  & 35.1, 63.2 & 5.0, 52.1  & 15.6, 57.5 & 8.4, 51.7  & 1.0, 49.8  & 6.2, 51.4  & 2.9, 49.1  & 16.3, 59.9 & 1.5, 49.6  & 2.5, 50.9  & 8.5, 52.8  & 148.1 \\
HSNet*        & 42.8, 68.4 & 10.0, 54.5 & 28.9, 60.0 & \textcolor{gray}{5.0, 52.1}  & 4.8, 51.0  & \textcolor{gray}{8.4, 51.7}  & 1.5, 50.9  & 4.9, 52.3  & 0.0, 49.7  & 6.3, 52.4  & 0.1, 49.7  & 0.0, 50.0  & 9.6, 53.6  & 148.1 \\
PATNet        & 51.1, 74.6 & 19.1, 58.0 & 50.3, 71.1 & 8.4, 53.9  & 21.3, 60.4 & 27.7, 66.4 & 1.0, 48.8  & 6.2, 51.3  & 2.0, 50.0  & 17.7, 64.5 & 2.0, 50.3  & 0.5, 50.1  & 19.2, 58.3 & 95.0  \\
PATNet*       & 57.3, 77.1 & 25.1, 62.1 & 53.9, 73.5 & \textcolor{gray}{8.4, 53.9}  & 16.2, 57.0 & \textcolor{gray}{27.7, 66.4} & 8.4, 55.4  & 10.6, 55.6 & 0.0, 49.9  & 6.5, 52.4  & 4.2, 51.8  & 0.0, 50.0  & 20.1, 58.8 & 95.0  \\
PerSAM        & 39.1, 52.4 & 1.3, 18.3  & 15.4, 48.6 & 0.0, 9.2   & 3.7, 41.2  & 10.1, 33.7 & 0.0, 35.3  & 5.4, 46.4  & 1.1, 38.9  & 1.4, 31.0  & 4.4, 51.3  & 1.0, 48.4  & 8.0, 37.9  & 5.2   \\
Matcher       & 14.0, 39.4 & 1.8, 46.7  & 32.6, 54.8 & 0.0, 18.7  & 6.2, 42.9  & 2.1, 14.5  & 0.0, 32.3  & 3.0, 39.8  & 1.1, 41.9  & 1.9, 35.3  & 2.0, 49.5  & 0.0, 46.7  & 5.0, 38.5  & 4.5   \\ \hline
SAM2-l(1024)  & 71.8, 84.1 & \textbf{48.1, 73.8} & 64.1, 80.1 & \textbf{58.2, 78.9} & 33.2, 65.9 & 53.5, 76.6 & 42.5, 71.2 & 51.3, 75.4 & \textbf{62.7, 81.3} & 43.1, 71.4 & 7.7, 53.4  & 20.1, 60.0 & 43.8, 72.7 & 14.1  \\
SAM2-b(1024)  & 76.6, 86.9 & 32.4, 65.7 & 71.7, 84.3 & 57.1, 78.4 & 32.4, 65.4 & 61.7, 80.7 & 45.8, 72.8 & 49.9, 74.7 & 46.2, 73.0 & 53.6, 76.6 & 10.5, 54.8 & 21.4, 60.6 & 45.0, 72.8 & 23.8  \\
SAM2-t(1024)  & 79.0, 88.4 & 31.3, 65.2 & 74.3, 85.7 & 55.3, 77.5 & 37.0, \textbf{67.8} & 60.0, 79.9 & 36.9, 68.4 & \textbf{57.8, 78.7} & 50.9, 75.4 & \textbf{56.3, 78.0} & 10.2, 54.7 & 14.1, 57.0 & 45.1, 73.1 & 34.5  \\
SAM2-s(1024)  & \textbf{80.5, 89.2} & 35.6, 67.4 & \textbf{75.8, 86.5} & 58.0, \textbf{78.9} & 33.6, 66.0 & \textbf{63.4, 81.5} & \textbf{50.8, 75.3} & 49.0, 74.2 & 55.5, 77.7 & 47.2, 73.4 & 15.4, 57.2 & \textbf{29.2, 64.6} & \textbf{47.9, 74.3} & 32.3  \\
SAM2-s(512)   & 72.2, 84.2 & 32.2, 65.7 & 72.9, 85.0 & 43.6, 71.6 & 28.3, 63.3 & 52.8, 76.2 & 42.6, 71.3 & 41.7, 70.5 & 28.6, 64.2 & 37.5, 68.6 & 4.6, 51.8  & 16.0, 57.9 & 38.8, 69.2 & 75.1  \\
SAM2-s(256)   & 62.4, 78.5 & 18.5, 58.7 & 67.1, 81.6 & 16.1, 57.8 & 15.9, 56.7 & 37.2, 68.4 & 5.9, 52.9  & 17.5, 57.9 & 7.6, 53.6  & 19.6, 59.4 & 4.1, 51.6  & 1.0, 50.2  & 23.9, 60.6 & 76.9  \\
FM-l           & 65.2, 80.7 & 14.9, 56.1 & 69.0, 82.3 & 22.4, 60.8 & 35.4, 66.7 & 50.3, 74.9 & 18.2, 59.0 & 39.9, 69.7 & 33.7, 66.8 & 33.5, 66.3 & 14.0, 56.7 & 22.9, 61.4 & 36.5, 66.8 & 277.8 \\
FM-s           & 62.3, 79.2 & 15.6, 56.4 & 68.1, 81.9 & 21.8, 60.4 & 33.1, 65.5 & 45.9, 72.7 & 17.9, 58.9 & 35.8, 67.7 & 37.3, 68.6 & 31.9, 65.5 & 10.6, 55.0 & 24.0, 62.0 & 34.6, 66.1 & \textbf{588.2} \\
FM-l+FastSAM-s & 70.1, 83.4 & 15.2, 56.0 & 69.1, 82.4 & 22.4, 60.8 & 36.6, 67.2 & 51.9, 75.8 & 18.2, 59.0 & 38.1, 68.7 & 35.6, 67.7 & 37.2, 68.2 & 19.7, 59.6 & 23.5, 61.7 & 38.2, 67.5 & 104.2 \\
FM-l+FastSAM   & 74.9, 86.1 & 17.2, 57.3 & 73.6, 84.9 & 22.4, 60.8 & \textbf{37.1}, 67.5 & 52.9, 76.2 & 18.2, 59.0 & 38.1, 68.7 & 36.0, 67.9 & 38.9, 69.1 & 19.4, 59.5 & 23.2, 61.5 & 39.3, 68.2 & 90.9  \\ \hline
\end{tabular}}
\label{CrossInResults}
\end{table*}

\begin{figure*}[htbp]
    \centering
		\includegraphics[width=\linewidth]{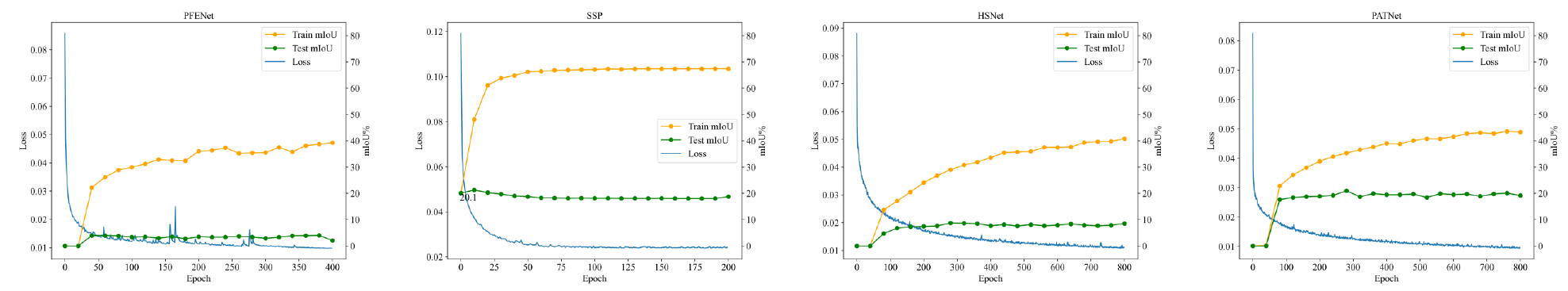}
	\caption{Details of the training processes for the selected methods in the cross-domain setting. In general, during training, the loss (\textcolor{blue}{blue}) of these methods decreases, and the mIoU on the training set (\textcolor{orange}{orange}) continuously improves. However, the mIoU on the test set (\textcolor{green}{green}) quickly saturates and remains relatively low.}
	\label{metalearning_process}
\end{figure*}

\subsection{Experimental results}
Table. \ref{CrossInResults} presents the quantitative results. For the meta-learning-based methods, it can be observed that these methods generally fail under the cross-domain setting, with the best mIoU and FB-IoU performance reaching only 19.2 and 58.3 respectively, achieved by the cross-domain method PATNet. This demonstrates that different industrial products typically involve domain-shift issues. Fig. \ref{metalearning_process} shows their detailed training processes. During training, these methods can generally fit the training set, reflected in the continuously decreasing loss and increasing mIoU of the training set. However, their generalization ability to the test set is poor, reflected in the rapid saturation of the test set mIoU and its relatively low values. It is worth noting that SSP is the only decoder-free method among these approaches, achieving 20.1 average mIoU on the entire dataset at epoch 0 before training. This performance surpasses the results of all meta-learning-based methods trained under cross-domain conditions, indicating that unsuitable meta-learning approaches even perform worse than directly utilizing the original pre-trained model without tuning. When using in-domain evaluation, the overall performance of these methods improves. However, there are also cases of poor performance in some object-based products, which may be attributed to the overfitting caused by the limited training set. Meanwhile, the in-domain setting requires a certain amount of defect samples, which are often difficult to obtain in real industrial environments, limiting its applicability. For existing VFMs-based FSS methods PerSAM and Matcher, they perform poorly in most industrial datasets. This may be due to the fact that these methods are primarily designed for natural images, which generally have distinct characteristics from industrial images. 

\begin{figure*}[t]
    \centering
		\includegraphics[width=0.9\linewidth]{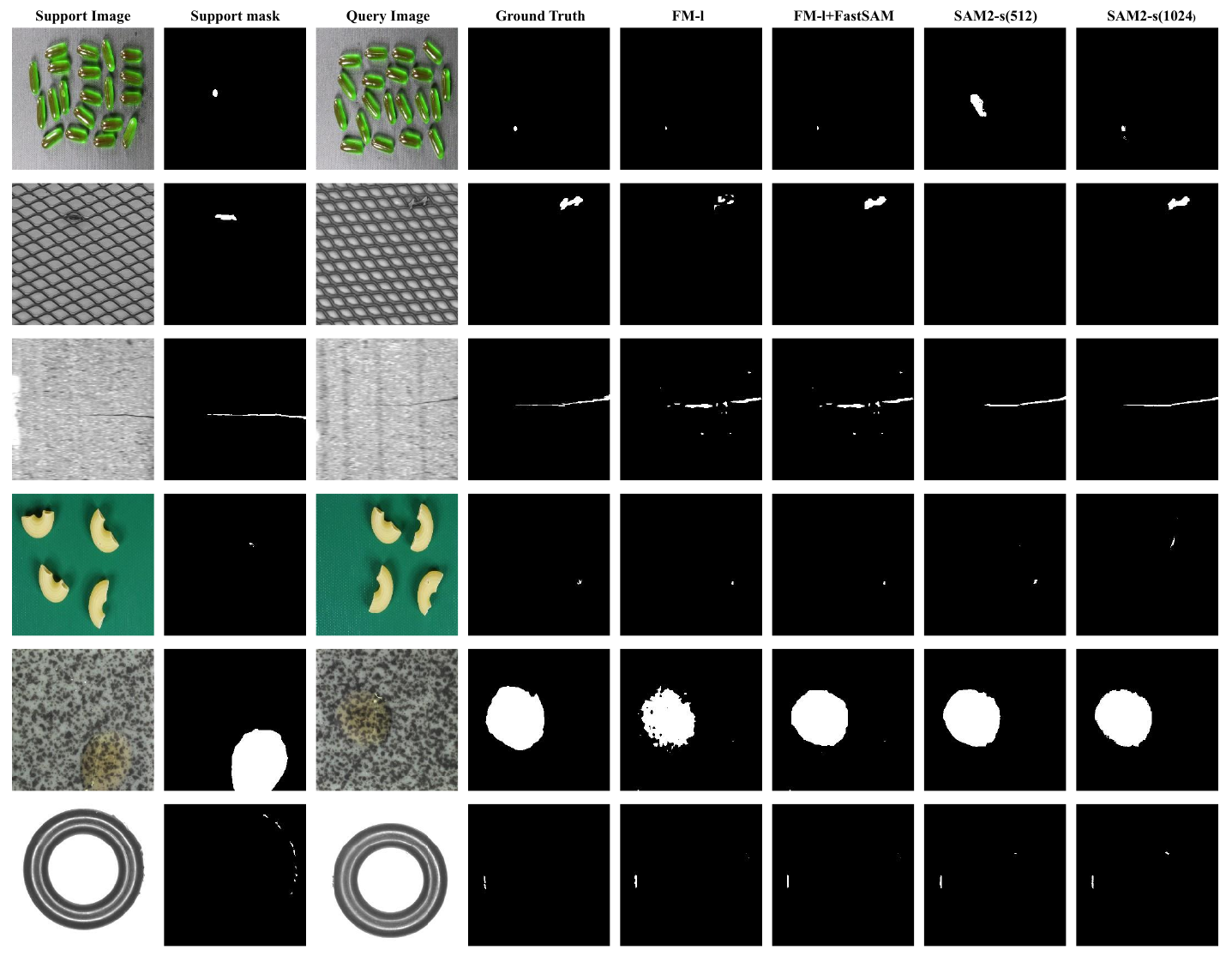}
	\caption{Qualitative results of our explored VFMs-based methods.}
	\label{quaresult}
\end{figure*}

In terms of the VFMs explored in our work, we observe that SAM2 demonstrates the strongest FDS performance. However, counterintuitively, a larger model does not always yield better performance. In our benchmark, the best performance is achieved by SAM2-s. Additionally, the image resolution setting has a significant impact on SAM2's performance. Upsampling the image to $1024\times1024$ can significantly improve the results. Our proposed feature-matching method demonstrates higher inference efficiency. Also, it outperforms existing meta-learning-based methods, including those trained with the in-domain setting. Fig. \ref{quaresult} presents some qualitative results.

Besides, from the perspective of the dataset, we observe that these methods generally perform better on texture-based products such as `Tile' and `Steel', but struggle with object-based products, such as our proposed rubber ring datasets and `capsules' and `macaroni2', even under the in-domain setting for those meta-learning-based methods. This demonstrates the importance of exploring FDS research in more comprehensive industrial applications rather than just for textures and also highlights our contribution regarding the proposed dataset.

\section{Discussion and ablation studies}
\label{Sec_Discussion}
\subsection{Analysis of Feature Matching-based FDS}
\subsubsection{Feature extractors}
We conduct experiments to investigate the performance of different pre-trained feature extractors on FDS. The selected models include DINO and DINOv2 series models and ImageNet-supervised pre-trained ResNet models. Specifically, for ResNet-based models, we explore their different intermediate layer features, including `layer2' and `layer3', as suggested in \cite{cohen2020sub}. For DINOv2, we upsample the image resolution to $266*266$ to ensure it is divisible by its patch size. The results are shown in Table. \ref{Table_extractor}. It can be observed that ViT models with small patch sizes and features from shallow layers in ResNet models generally perform better, indicating that high-resolution features are crucial for the FDS task. Among the existing models, the DINO and DINOv2 ViT models with small patch sizes perform the best, but their inference speeds are relatively slow. In comparison, our distilled model is not only faster but also surpasses its original teacher model in performance.

\subsubsection{Feature distillation}
\begin{figure*}[htbp]
    \centering
		\includegraphics[width=\linewidth]{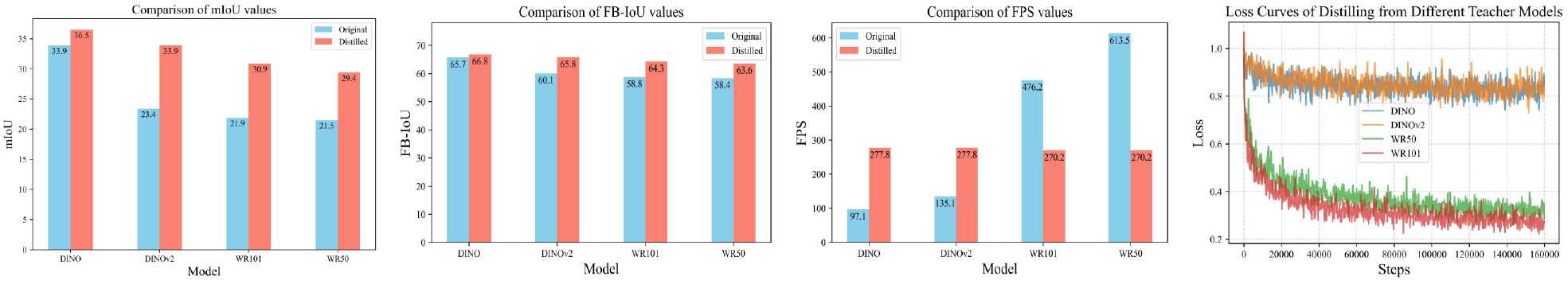}
	\caption{Comparison of results with different teacher models and their corresponding loss curves. `WR101' and `WR50' are the abbreviations of `WideResNet-101' and `WideResNet-50' respectively. For these ResNet models, we use their features from `layer2'. We use the ViT-S/8 model for DINO and ViT-S/14 for DINOv2 respectively. `FPS' refers to `frames per second'}
	\label{disprocess}
\end{figure*}

We further evaluate the impact of different teacher models on the results. Specifically, we conduct experiments using the `layer2' of the ImageNet supervised ResNet models and the DINOv2 ViT-S/14 as teacher models, while keeping the student model architecture as the default of FM-l. In particular, to align the student's output with these teacher models, for the selected DINOv2 model, we modify the input image size of the teacher model to $896 \times 896$ and for the ResNet models, we modify the student model's output channel to 512. Fig. \ref{disprocess} presents a comparison of the results from different teacher models. In general, the segmentation performance of all the distilled student models exceeds that of the original teacher models. This suggests that distilling high-resolution features can generally benefit FDS. Meanwhile, the higher the performance of the teacher models, the better the performance of the distilled student models. From the comparison of loss curves, we observe that the student model fits the ViT-based DINO models less effectively than the CNN-based ResNet models. However, distillation from the DINO series models yields better results, and the inference speed is significantly improved compared to the original ViT models.

\begin{table*}[htbp]
\centering
\caption{Quantitative comparison of the proposed method with different feature extractors and with (w/) and without (w/o) using FastSAM for refinement. `R50', `WR101', `WR50', and `R18' are the abbreviations of `ResNet50', `WideResNet-101', `WideResNet-50', and `ResNet-18' respectively. `l2' and `l3' are the abbreviations of `layer2' and `layer3'. `FPS' refers to `frames per second'.}

\label{Table_extractor}
\scalebox{0.6}{
\begin{tabular}{c|cccccc|ccc|cccccc|cc}
\hline
\multirow{2}{*}{} & \multicolumn{6}{c|}{DINO}                                 & \multicolumn{3}{c|}{DINOv2}    & \multicolumn{6}{c|}{ImageNet-supervised}                  & \multicolumn{2}{c}{Ours} \\ \cline{2-18} 
                  & ViT-S/16 & ViT-S/8 & ViT-B/16 & ViT-B/8 & R50-l2 & R50-l3 & ViT-S/14 & ViT-B/14 & ViT-L/14 & WR101-l2 & WR101-l3 & WR50-l2 & WR50-l3 & R18-l2 & R18-l3 & FM-l         & FM-s        \\ \hline
mIoU              & 21.6     & 33.9    & 19.5     & 34.7    & 17.5   & 10.5   & 23.4     & 23.2     & 22.9     & 21.9     & 9.9      & 21.5    & 11.9    & 19.7   & 14.2   & \textbf{36.5}        & 34.6       \\
FB-IoU            & 58.7     & 65.7    & 57.5     & 65.9    & 55.6   & 47.8   & 60.1     & 59.9     & 59.7     & 58.8     & 46.3     & 58.4    & 50.1    & 58.1   & 53.9   & \textbf{66.8}        & 66.1       \\
FPS               & 142.8    & 97.1    & 128.2    & 47.6    & 714.2  & 357.1  & 135.1    & 117.6    & 58.8     & 476.2    & 204.1    & 613.5   & 349.2   & 833.3  & 689.7  & 277.8       & \textbf{588.2}      \\ \hline
\end{tabular}}
\end{table*}

\subsubsection{Prototype strategy}
During feature matching, we apply patch-level aggregation for foreground feature vectors to build foreground prototypes, while preserving the original dense background feature vectors as background prototypes. Here, we evaluate the performance of different prototype strategies, including using the original dense feature vectors, applying patch aggregation, and employing global average pooling, abbreviated as `dense', `patch', and `pool' respectively. We use $f_{strategy}$ and $b_{strategy}$ to denote the foreground and background prototypes with different aggregation strategies. For example, $f_{pool}$ represents applying global average pooling to build foreground prototypes. Table. \ref{Table_prototype} presents the results. It can be observed that global pooling is generally not suitable in this case, especially for the background vectors. The use of patch aggregation can improve the performance. When using FastSAM for refinement, the optimal performance is achieved by applying patch aggregation solely to the foreground features. Without FastSAM, the best performance is obtained by applying patch aggregation to both the foreground and background features.

\begin{table}[htbp]
\centering
\caption{Quantitative comparison of FM-l with different prototypes.}
\label{Table_prototype}
\scalebox{0.7}{
\begin{tabular}{c|cc|cc}
\hline
\multirow{2}{*}{Prototypes} & \multicolumn{2}{c|}{w/ FastSAM} & \multicolumn{2}{c}{w/o FastSAM} \\ \cline{2-5} 
                            & mIoU          & FB-IoU          & mIoU          & FB-IoU          \\ \hline
$f_{patch}, b_{dense}$      & \textbf{39.3}          & \textbf{68.2}            & 36.5          & 66.8            \\
$f_{patch}, b_{patch}$      & 38.4          & 67.4            & \textbf{37.1}          & \textbf{67.0 }           \\
$f_{patch}, b_{pooling}$    & 7.2           & 38.5            & 7.6           & 39.3            \\
$f_{dense}, b_{dense}$      & 37.9          & 67.3            & 36.2          & 66.6            \\
$f_{pooling}, b_{dense}$    & 29.0          & 63.5            & 24.8          & 61.1            \\ \hline
\end{tabular}}
\end{table}

\subsubsection{Fusion strategy with FastSAM}
Our method introduces a specialized fusion algorithm to fuse the feature matching and FastSAM results. To verify its effectiveness, we here compare it with two alternative fusion strategies, including (1) using a simple union operation to fuse the results, denoted as $R_0 \cup R_{sam}$ and (2) only adopting the selected masks of FastSAM, denoted as $R_{sam}$. The results are shown in Table. \ref{Table_fusion}. It can be observed that our fusion strategy achieves the optimal performance here.

\begin{table}[htbp]
\centering
\caption{Quantitative comparison of the proposed method with different fusion strategies. `FPS' refers to `frames per second'}
\label{Table_fusion}
\scalebox{0.7}{
\begin{tabular}{c|cccc}
\hline
Fusion Strategy       & mIoU & FB-IoU   \\ \hline
FM-l                   & 36.5 & 66.8    \\
$R_{sam}$             & 22.1 & 59.7  \\
$R_0 \cup R_{sam}$    & 37.8 & 67.4   \\
our fusion strategy   & \textbf{39.3} & \textbf{68.2}    \\ \hline
\end{tabular}}
\end{table}

\subsection{Analysis of SAM2-based FDS}
In Sec. \ref{Exploring SAM2}, we illustrate how to leverage SAM2's video track mode to implement FDS. Intuitively, SAM2 can also be used, like FastSAM, to refine the feature-matching results. Here we conduct experiments to validate this approach, without changing our original fusion strategy for $R_0$ and $R_{sam}$, but only replacing FastSAM with SAM2.
Different from FastSAM, which can directly generate all the potential masks without considering the prompts, SAM2 requires the encoded prompt features to achieve segmentation. 
More detailed prompts typically lead to more accurate segmentation results, but the inference time also increases. Here, we evaluate various prompt strategies, including: (1). \textbf{Everything mode}: this mode divides the entire image into evenly spaced grid points and uses these points as prompts. We test different sampling rates, including sampling the image into $64 \times 64$ points, referred to as `everything(64)', and $32 \times 32$ points, referred to as `everything(32)'; (2). \textbf{Point mode}: we convert the feature matching result $R_0$ into points as input. Additionally, we explore evenly spaced sampling of these points to reduce the inference cost. We test 8x downsampling. These strategies are referred to as `points' and `points(8x)', respectively. We also experiment with the center points, i.e., the center of each morphologically disconnected region in $R_0$, referred to as `center points'. (2). \textbf{Box mode}: we extract the minimum bounding rectangle for each morphologically disconnected region in $R_0$, referred to as `boxes'. 
In addition to these prompt configurations, we use the default configuration of the official SAM2 and choose the SAM2-s model, with the image upsampled to $1024 \times 1024$. The results are presented in Table. \ref{sam2results}. 
Under dense prompts like `everything(64)', SAM2 achieves better segmentation results than FastSAM, but the inference speed is significantly reduced. Meanwhile, the selected prompt strategies consistently fall short of directly using SAM2's video track mode both in terms of efficiency and accuracy. 

\begin{table*}[htbp]
\centering
\caption{Quantitative comparison of using SAM2 with different prompt strategies to refine the feature matching results. `SAM2-s(1024)' refers to leverage the video track mode of SAM2.}
\scalebox{0.7}{
\begin{tabular}{c|ccccccc|cc}
\hline
       & everything(64) & everything(32) & points & points(8) & center points & boxes & center points+boxes & SAM2-s(1024)  & FM-l+FastSAM   \\ \hline
mIoU   & 40.2           & 39.3           & 37.1   & 36.7      & 36.7          & 36.5  & 36.8                & \textbf{47.9} & 39.3          \\
FB-IoU & 68.3           & 67.9           & 66.9   & 66.6      & 66.7          & 66.8  & 66.7                & \textbf{74.3} & 68.2          \\
FPS    & 0.2            & 0.7            & 0.8    & 5.2       & 53.6          & 53.6  & 23.8                & 32.3          & \textbf{90.9} \\ \hline
\end{tabular}}
\label{sam2results}
\end{table*}

\section{Conclusion}
    This paper focuses on the FDS task in general industrial scenarios, conducting a comprehensive exploration from the perspective of metric learning. At the dataset level, we observe that existing benchmarks generally contain scarce object-based products. To address this, we contribute a new real-world dataset. We also reorganize some existing datasets to build a more comprehensive FDS benchmark. Through a detailed evaluation of meta-learning-based methods and VFMs, we emphasize the necessity of expanding the scope of FDS evaluation beyond just simple textures. From the methodological perspective, we observe that existing meta-learning approaches are generally unsuitable for this task as they heavily rely on in-domain training data, which is generally difficult to obtain in real industrial settings. Regarding VFMs, on the one hand, we explore an efficient route based on feature matching and identify the importance of high-resolution features in this context. We propose a new feature-matching method, which includes knowledge distillation, prototype construction, and refinement using FastSAM. Overall, it achieves convincing performance while maintaining higher inference efficiency. In addition, we explore the SAM2-based approach and find that its video track mode demonstrates strong performance for FDS. We believe that with the advent of more powerful feature extractors and large segmentation models in the future, the techniques we have explored will continue to improve.

\section*{Acknowledgement}
The calculations were performed by using the HPC Platform at Xi’an Jiaotong University. This work was supported by the Funding of Xi'an Aerospace Propulsion Institute and the National Natural Science Foundation of China (grant number 52293405).

\bibliographystyle{elsarticle-num} 
\bibliography{ref}

\end{document}